\documentclass{article}

\PassOptionsToPackage{numbers, compress}{natbib}

\usepackage[preprint]{neurips_2024}

\usepackage[utf8]{inputenc} 
\usepackage[T1]{fontenc}    
\usepackage{hyperref}       
\usepackage{url}            
\usepackage{booktabs}       
\usepackage{amsfonts}       
\usepackage{nicefrac}       
\usepackage{microtype}      
\usepackage{xcolor}         

\usepackage{multicol}
\usepackage{multirow}
\usepackage{adjustbox}
\usepackage{amsmath}

\usepackage{wrapfig}

\usepackage{listings}
\title{Towards Data-Centric Automatic R\&D}

\author {
    Haotian Chen\textsuperscript{\rm 1}\thanks{Equally Contributed} , Xinjie Shen\textsuperscript{\rm 1}\footnotemark[1] , Zeqi Ye\textsuperscript{\rm 1} , Wenjun Feng\textsuperscript{\rm 1}, Haoxue Wang\textsuperscript{\rm 1} \\
    \textbf{Xiao Yang\textsuperscript{\rm 2}\thanks{Corresponding Author} , Xu Yang\textsuperscript{\rm 2}, Weiqing Liu\textsuperscript{\rm 2}, Jiang Bian\textsuperscript{\rm 2}} \\
    \textsuperscript{\rm 2}Microsoft Research Asia \\
    \texttt{\{ht1ian.chen, frinkleko, liamyzq\}@gmail.com} \\
    \texttt{fwj20020813@outlook.com}, \texttt{whx924@gmail.com} \\
    \texttt{\{xiao.yang, xuyang1, weiqing.liu, jiang.bian\}@microsoft.com}
}

\begin{document}

\maketitle

\begin{abstract}
    The progress of humanity is driven by those successful discoveries accompanied by countless failed experiments. Researchers often seek the potential research directions by reading and then verifying them through experiments. The process imposes a significant burden on researchers. In the past decade, the data-driven black-box deep learning method has demonstrated its effectiveness in a wide range of real-world scenarios, which exacerbates the experimental burden of researchers and thus renders the potential successful discoveries veiled. Therefore, automating such a research and development (R\&D) process is an urgent need. In this paper, we serve as the first effort to formalize the goal by proposing a \textbf{R}eal-world \textbf{D}ata-centric automatic \textbf{R}\&\textbf{D} \textbf{Bench}mark, namely $\text{RD}^2$Bench. $\text{RD}^2$Bench benchmarks all the operations in data-centric automatic R\&D (D-CARD) as a whole to navigate future work toward our goal directly. We focus on evaluating the interaction and synergistic effects of various model capabilities and aiding in selecting well-performing trustworthy models.
    Although $\text{RD}^2$Bench is very challenging to the state-of-the-art (SOTA) large language model (LLM) named GPT-4, indicating ample research opportunities and more research efforts, LLMs possess promising potential to bring more significant development to D-CARD: They are able to implement some simple methods without adopting any additional techniques. We appeal to future work to take developing techniques for tackling automatic R\&D into consideration, thus bringing the opportunities of the potential revolutionary upgrade to human productivity.
\end{abstract}

\section{Introduction}
\begin{quote}
    \noindent \textit{``I have not failed. I've just found 10,000 ways that won't work.''}
\end{quote}
\hfill --- \textit{Thomas Alva Edison}

\renewcommand{\thefootnote}{}
\footnotetext{This is an open-source project starting in Oct. 2023. \textsuperscript{\rm 1} Work done during an internship at Microsoft.}
\renewcommand{\thefootnote}{\arabic{footnote}}

\begin{figure}[h]
    \centering
    \includegraphics[width=1.0\textwidth]{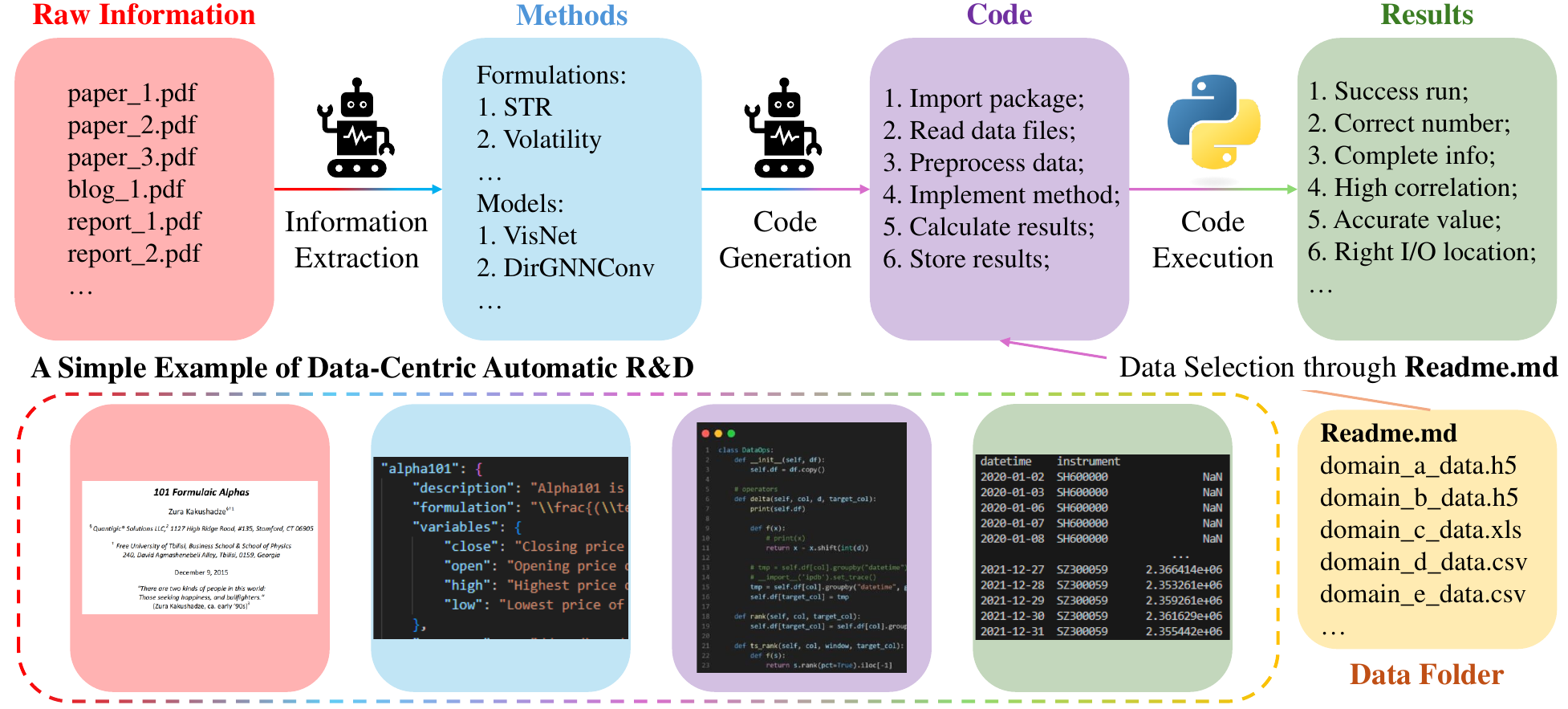}
    \caption{An overview of the R\&D process. Researchers read papers and reports to extract the implementable methods (usually formulated as mathematical formulas or model architectures) for seeking potential research directions. Then, they accurately implement the methods to obtain the results for further analysis and development. }
    \label{fig:overview}
\end{figure}

The advancement of human society and the enhancement of living standards are highly correlated with the development of technology~\citep{smith1937wealth,ranis1961theory,perez2003technological,brynjolfsson2014second}. Numerous truths and principles remain undiscovered in the world, awaiting experimental exploration~\citep{shapere1964structure,popper2005logic}. Those few successful discoveries, accompanied by countless failed experiments, propel the frontiers of technology. Historically, scientific researchers, including Edison, have undertaken extensive experiments by conducting them manually.
In the age of AI, the influence of data-driven solutions, such as machine learning (ML) systems, is rapidly expanding~\citep{mikolov2013distributed,devlin2018bert,openai2023gpt4}. These systems are known for their robust fitting capabilities and their ``black box'' nature, which significantly increases the experimental load on researchers and hinders the process of identifying and validating effective methodologies. This paper concentrates on this critical scenario, which we refer to as \textit{Data-Centric Research and Development (R\&D)}.
To cope with the prohibitively expensive costs and the overwhelming volume of experiments required, we consider automating such an R\&D process for higher research efficiency by leveraging the strong language understanding and programming ability of the state-of-the-art (SOTA) large language models (LLMs)~\citep{srivastava2023beyond}. The brief illustration of \textbf{D}ata-\textbf{C}entric \textbf{A}utomatic \textbf{R}\&\textbf{D} (D-CARD) is shown in Figure~\ref{fig:overview}.

The first step towards automatic R\&D is to formalize the task and provide a benchmark for identifying the potential effective methods and research directions. Intuitively, an outstanding methodology identified by the benchmark should possess (1) strong \textbf{language understanding ability} to identify the implementable methods or ideas (e.g., formulations and models) in the given raw information (e.g., papers, reports, websites, etc.) and (2) strong \textbf{implementation ability} to accurately implement the methods by programming and then obtain reliable experimental results. Previous work focuses on benchmarking the different aspects of the two abilities. Specifically, the language understanding ability of LLMs is partly evaluated through analyzing their performance on relation extraction~\citep{wadhwa2023revisiting}, question answering~\citep{zhuang2023toolqa}, and other natural language processing (NLP) tasks~\citep{qin2023chatgpt}. Meanwhile, the implementation ability of LLMs is partly tested through benchmarks like SWE-Bench~\citep{jimenez2023swe}, ToolBench~\citep{qin2023toolllm}, ML-Bench~\citep{liu2023mlbench} and MetaTool~\citep{anonymous2024metatool}, which study their ability of solving GitHub issues, using tools to program, and determining whether to use tools in a given scenario.

In this paper, we serve as the first effort to investigate the capabilities of the SOTA LLMs in tackling automatic R\&D and propose a \textbf{R}eal-world \textbf{D}ata-centric automatic \textbf{R}\&\textbf{D} \textbf{Bench}mark ($\text{RD}^2$Bench). The scenario studied by $\text{RD}^2$Bench possesses two unique and distinct characteristics that fundamentally differentiate it from others. First, $\text{RD}^2$Bench focuses on studying the real-world scenario where all the operations in R\&D are automatic and evaluated as a whole, thus navigating the related future research efforts toward the goal of developing human technology more effectively. The real-world scenario requires more comprehensive and advanced model capabilities and exhibits new challenges. Second, we study the real-world automatic R\&D in data-centric settings to navigate future work toward the urgent experimental exploration need brought by black-box data-driven models.
Compared with existing benchmarks, $\text{RD}^2$Bench possesses two significant advantages:


(1) \textbf{$\text{RD}^2$Bench evaluates the interaction and synergistic effects of various model capabilities} instead of focusing on a single aspect of ability, which not only captures the frontier of SOTA LLMs but also bridges the gap between studying ``individual ability'' and ``real-world synergistic effects of abilities''. In automatic R\&D, an ML system fails to complete the task even if it possesses both the strong information extraction ability and the strong programming or tool-using ability: While it succeeds in extracting methods and implementing them, it fails in selecting the appropriate data from the datasets or misunderstanding either the descriptions of data features or the requirements expressed by prompts. Additionally, exhaustively enumerating all the aspects for benchmarking is extremely challenging, which is overcome by $\text{RD}^2$Bench.

(2) \textbf{$\text{RD}^2$Bench tends to select well-performing trustworthy models} instead of those models that fail to learn rationales and causality yet possess outstanding performance. Specifically, ML systems easily achieve SOTA performance on previous benchmarks by shortcut learning or learning spurious correlations instead of learning rationales or causality~\citep{mudrakarta2018did,geirhos2020shortcut,cui2022stable,wang2022identifying,chen2023did}. This renders a benchmark ineffective and misleading as it fails to accurately identify the well-performing trustworthy methods. For example, an ML system achieves SOTA performance on dog classification by merely recognizing grass~\citep{zhang2021deep}. $\text{RD}^2$Bench, on the contrary, eliminates such models by its high difficulty and large scope. The decision rules of models have to simultaneously satisfy at least four major requirements: (1) accurately and comprehensively extracting the implementable methods; (2) precisely selecting the method-specific data for computation; (3) correctly writing the code according to the logic expressed by methods and prompts; (4) successfully storing the correct results in a predefined format. Therefore, the decision rules of models selected by this benchmark are stable (work well in various situations), and thus getting closer to rationales and causality~\citep{cui2022stable}.


We evaluate the existing SOTA LLMs on $\text{RD}^2$Bench to expose their bottleneck and characterize the future research direction. $\text{RD}^2$Bench reveals new insights: (1) Among the popular LLMs, GPT-4 exhibits promising potency in dealing with the D-CARD task; (2) Detailed information of data descriptions significantly improves the performance of GPT-4; (3) The ability to query domain-specific knowledge is a basic requirement of D-CARD methods; (4) The more complex the method is, the more unstable the model's performance is.

\section{Related Work}
\subsection{LLM as Autonomous Agent}
In the past few years, LLM has made great achievements in both academia and industry~\citep{GPT4Report,Llama2}, and has achieved results that surpass the previous level in a number of classic tasks~\citep{LLMSurveyOnClassicalNLP}. Research has shown that with the growth of data volume and model size~\citep{Emergent}, LLM has emerged with stronger reasoning and other capabilities~\citep{ouyang2022training}. These capabilities enable LLM to exhibit certain agent-like behaviors in some tasks such as using or creating tools~\citep{ToolLLM,toolCreate}, planning~\citep{TreeofT,FewShotLearner}, and memory. Therefore, more and more researchers have expressed their expectations for its human-like and overall capabilities, and have made preliminary explorations of it as an independent agent~\citep{Voyager,Reflexion}. Multi-agent collaboration~\citep{AutoGen,Li2023TheoryOM} is also introduced to LLM for better accuracy and generalizability. Moreover, for reducing human efforts and automatically exploring, previous work focuses on autonomous LLM agents for general purpose are purposed~\citep{AutoGPT,HuggingGPT}. Positive views further believe that the realization of AGI may come from the evolution of autonomous LLM and some inspiring examples have been released~\citep{BabyAGI}.

However, most research still focuses on limited scenarios that are given with clear and fixed questions and backgrounds. A recent work~\citep{RandD} has attempted to introduce LLM to the R\&D field and formalize the R\&D process as a sequence of tasks. However, there is no easy-to-use benchmark for the community and current R\&D tasks may be too general and can't reveal significant signals. In this work, we propose a benchmark for LLM in data-centric R\&D tasks and provide a comprehensive evaluation.

\subsection{Semi-Automatic R\&D with Agents}
Scientific research and development (R\&D) is a time-consuming and important process. In the past, R\&D has been mainly conducted by human researchers with countless failed experimental explorations and creative observation conclusions. Agents have been introduced to R\&D to reduce human efforts and automatically explore. Recently, there have been attempts to partly automate R\&D, including the automatic chemical synthesis planning ~\citep{AutoChem}, automatic molecular design ~\citep{autoMo,autoDrug,AutoChem}, automatic theorem proving ~\citep{DT-Solver,leandojo}. However, these attempts mainly focus on automatic searching for possible solutions and optimizations with symbolic representation ~\citep{symbolic} and heuristic techniques ~\citep{TheoremSearch}, but less addressing long-horizon planning, implementation, and reasoning for the next step idea exploration. Moreover, the data-centric R\&D tasks currently have not been explored in the community, and no benchmark has been proposed for the community. Previous works have applied LLM to real-world R\&D tasks such as debugging issues ~\citep{DebugBench,githubIssue} or only focus on data-centric but not real-world R\&D tasks ~\citep{dataCentericButnoRD}. In this work, we propose a benchmark for LLMs in data-centric R\&D tasks and evaluate the performance of LLMs.


\section{$\text{RD}^2$Bench}
Overall, our benchmark focuses on evaluating the finally implemented results according to the given raw information (e.g., papers, reports, websites, etc.). Moreover, we also provide human-annotated ground-truth information corresponding to the intermediate steps for debugging and more comprehensive evaluation. $\text{RD}^2$Bench selects well-performing models that follow human operations and accurately calculate the final results. We introduce the details of our proposed $\text{RD}^2$Bench in the following sections. In section~\ref{sec:data collection} and section~\ref{sec:annotation}, we introduce how we collect data and perform human annotation to form $\text{RD}^2$Bench. Then, we elaborate on the two necessary steps, namely method extraction and method implementation, to perform R\&D in section~\ref{sec:extraction} and section~\ref{sec:implementation}. Finally, we detail our adopted evaluation metrics in section~\ref{sec:metrics}.

%
\subsection{Data Collection}
\label{sec:data collection}
We consider the raw information that contains formulas and models, which represent a wide range of methods proposed in the AI domain. 

\noindent \textbf{Data Collection with Formulas.} 
We prepare raw information that contains formulas as the input of R\&D. Raw information is presented as publicly available financial reports and stock trading data. 
Formulas are usually mathematical formulas that take complex numeric input data about stock, company, and market as input and output a series of values with the time series.
We collect financial reports with 27 implementable formulas distributed in three difficulty levels: easy, medium, and hard. Domain experts manually label the difficulty levels according to the complexity of implementation. To obtain their implementation results, an agent is expected to accurately select the features from three types of trading data scattered across 2010 to 2022, namely fundamental, price-volume, and high-frequency data. We denote the three types of data as Data I, II, and III, respectively. 

\noindent \textbf{Data Collection with Models.} 
We collect papers with six open-sourced models~\citep{antisymmetric,DirGNN,GPSConv,linkx,pmlp,Visnet}. The implementation of models adopts Pytorch~\citep{pytorch} and torch\_gemometric framework~\citep{pyg} to perform deep learning. All the papers and models are publicly available. We manually label the difficulty level (easy, medium, hard) of the task based on the complexity of implementation (computational graphs and tensor operations). We refer the readers to the appendix for more details about the dataset and the task.


\subsection{Human Annotation.}
\label{sec:annotation}
To provide a more comprehensive evaluation for debugging and analyzing, we conduct human annotation to provide the ground-truth results of our collected data, namely method extraction results and method implementation results.

\noindent \textbf{Challenges.}
We confront five main challenges in the human annotation process.
First, we need to identify the difficulty levels of methods to ensure the diversity of our benchmark and expose the bottleneck of current models. 
Second, we have to identify and discard the raw information if its presented methods demand unavailable data: The computation of some formulas can require confidential information that is not publicly available. 
Third, since the definitions or descriptions of some methods can be vague, leading them to be unimplementable, we have to filter out these methods. 
Fourth, some domain-specific methods containing factual errors should be filtered out since they are not implementable. 
Fifth, we should distinguish the domains and types of the methods according to their descriptions.
To sum up, all the challenges imply the fact that human annotation of $\text{RD}^2$Bench requires high time cost and expertise of annotators. Therefore, we commit more effort to designing the annotation guidelines and mechanisms to ensure the dataset quality. 

\noindent \textbf{Annotation Guidelines and Mechanisms.}
We provide principled guidelines and mechanisms to the annotators. Specifically, we first clarify our requirements in the guidelines: The extracted ground-truth methods are implementable without the requirement of any additional information, and the ground-truth implementation results are stored in an expected space with the predefined form and shape. Second, we provide training to the annotators. After the training step, we delicately design a test case and an interview to examine if the annotators possess expertise in the target domain (e.g., financial domain) and understand the guidelines. Third, we make each annotation result checked at least twice by annotators. A senior researcher and an applied scientist with domain expertise and rich engineering experience will further check each annotation result together before it finally gets included in the benchmark.

\subsection{Method Extraction Step}
\label{sec:extraction}
We evaluate the ability of models to recognize and extract information from raw information (e.g., R\&D context). A qualified model is expected to discern feasible methods (formulas and models) from extensive research materials and extract all necessary information for implementing these formulas. The ability serves as the foundational premise for subsequent code implementation. 

We expect the model to accurately and comprehensively extract the methods mentioned in the research materials it reviews, including all essential conditions required to implement the method. For methods with incomplete information, further implementation is not required; for methods with complete conditions, a model is expected to correctly comprehend the semantic meaning of these conditions stated in natural language and generate corresponding code. Specifically, we have predefined the extraction format (key-value pair) for the model. We employ the F1 score to measure the comprehensiveness and accuracy of method identification and extraction. 

Note that some methods in the original materials might only imply their function, effect, or origin through their names without explicitly presenting their formulas, definitions, or other details. In such cases, the model may choose not to extract them or opt to autonomously complete them based on the semantics of the original materials. We expect the latter approach in future work, as it showcases the creativity of models by proposing new formulas and generating brand-new, informative, and reliable information. In the current version of the benchmark, only methods mentioned by name are evaluated in this manner; future iterations will explore and assess the model's ability to generate new names and formulas when none are explicitly mentioned.

\subsection{Method Implementation Step}
\label{sec:implementation}
\begin{figure}[h]
    \centering
    \includegraphics[width=\textwidth]{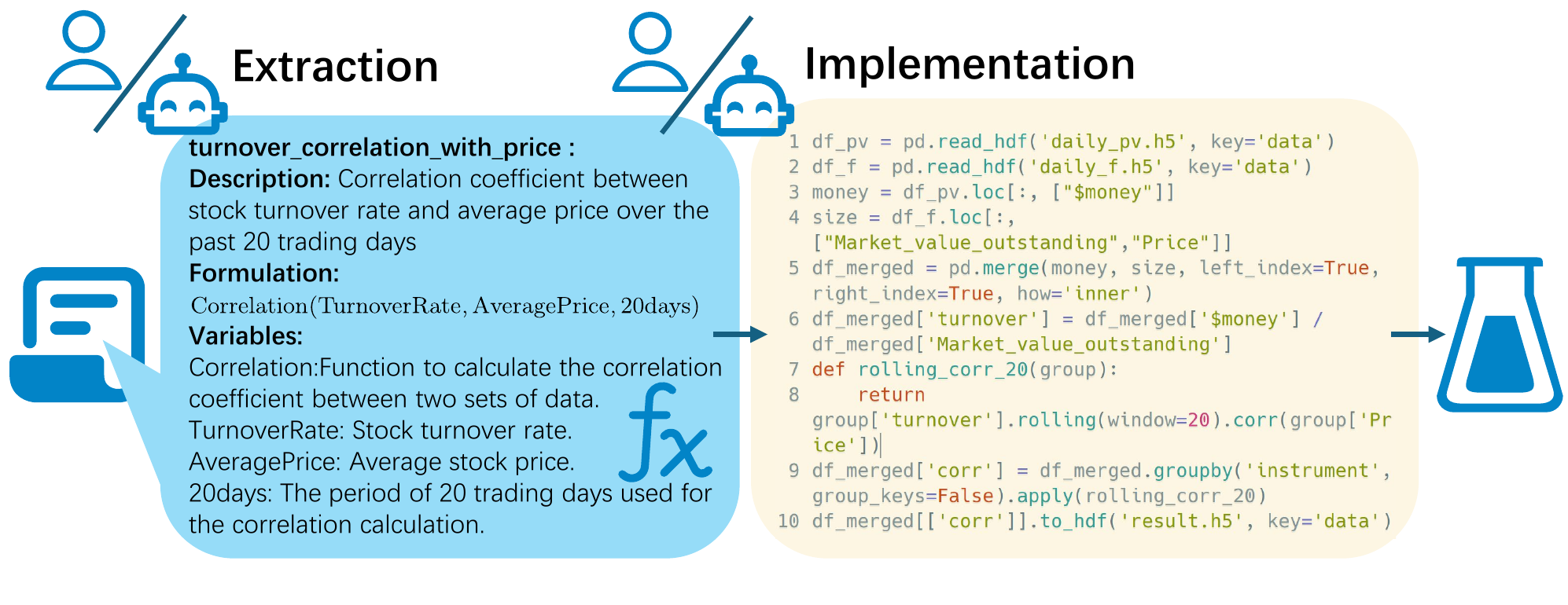}
    \caption{An example of a formula implementation task.}
    \label{fig:formula_implementation_example}
\end{figure}

In this section, we evaluate the performance of LLM in the implementation of methods. Given all the necessary conditions provided to the model after the previous step, the model needs to select the necessary data and write code from scratch to implement the method with an informative and well-organized prompt. Details of the prompt are included in the dataset, which is also shown in the appendix. We encourage models to use Python and perform data analysis. They are also permitted to use common machine-learning libraries.
One example of the method implementation step is shown in Figure~\ref{fig:formula_implementation_example}.

\subsection{Evaluation Metrics}
\label{sec:metrics}
We adopt multiple metrics to evaluate model performance in each step. For formula implementation, we adopt the average and maxima ``running success rate'', ``format success rate'', ``Pearson correlation'' and ``value accuracy'' across multiple independent attempts. We use ``avg.'', ``exe.'', ``form.'', ``corr.'', and ``acc.'' to denote the average value, number of successful execution times, number of matched result formats, the correlation, and the accuracy of corresponding values, respectively. We refer the readers to more details about the metrics calculation details in App~\ref{App:A}.

For model implementation, we believe a successful implementation of a model should be consistent with the ground truth implementation as the model can be viewed as a numeric function and combination of tensor transformations. Therefore, we propose these two metrics for the model architecture implementation task: tensor shape consistency rate (tsc.), tensor value consistency rate (tvc.). Specifically, for each model layer, we calculate the consistency rate of the tensor shape and tensor value between the ground truth implementation and the implementation generated by the LLM. All the ground truth tensor values are determined by ground truth implementation codes with random Gaussian noise.
Therefore, the formula for the two metrics is as follows, where $S_{\text{shape}}^i$ and $S_{\text{value}}^i$ are the consistency rate of tensor shape and tensor value in layer $i$, respectively, and $d_i$ is the maximum length of the two tensors as the two tensors are $\mathbf{Z}_{i}$ and $\mathbf{Z}_{i}^{*}$, the ground truth and the generated tensor, respectively:
{\small
\begin{equation}
    \begin{gathered}
        S_{\text{shape}}^i(\mathbf{Z}_{i}, \mathbf{Z}_{i}^{*})  = \left(1+\exp\left(\frac{\sum_{j=1}^{d}|\text{dim}(\mathbf{Z}_{i})_{j} - \text{dim}(\mathbf{Z}_{i}^{*})_{j}|}{d}\right)\right)^{-1}, \\
        S_{\text{value}}^i(\mathbf{Z}_{i}, \mathbf{Z}_{i}^{*})  = \left(1+\exp\left(\frac{\sum_{j=1}^{d}|\mathbf{Z}_{i}^{(j)} - \mathbf{Z}_{i}^{*(j)}|}{d}\right)\right)^{-1}, d = \max(\text{len}(\text{dim}(\mathbf{Z}_{i})), \text{len}(\text{dim}(\mathbf{Z}_{i}^{*}))),
    \end{gathered}
\end{equation}
}
while the shorter tensor is padded with zeros to match the length of the longer tensor.
As the final score of the two metrics, we use the weighted sum of the consistency rate of all layers, weight increases with the depth of the layer and is summed as one: $S_{\text{final}} = \frac{\sum_{i=1}^{n}S^i \cdot \gamma^i}{\sum_{i=1}^{n}\gamma^i},$
where $n$ is the number of layers in the model, $\gamma$ is a tunable hyperparameter to control the weight increase, and we set $\gamma = 1.1$ in our experiments.

\begin{figure}[h]
    \centering
    \includegraphics[width=0.8\textwidth]{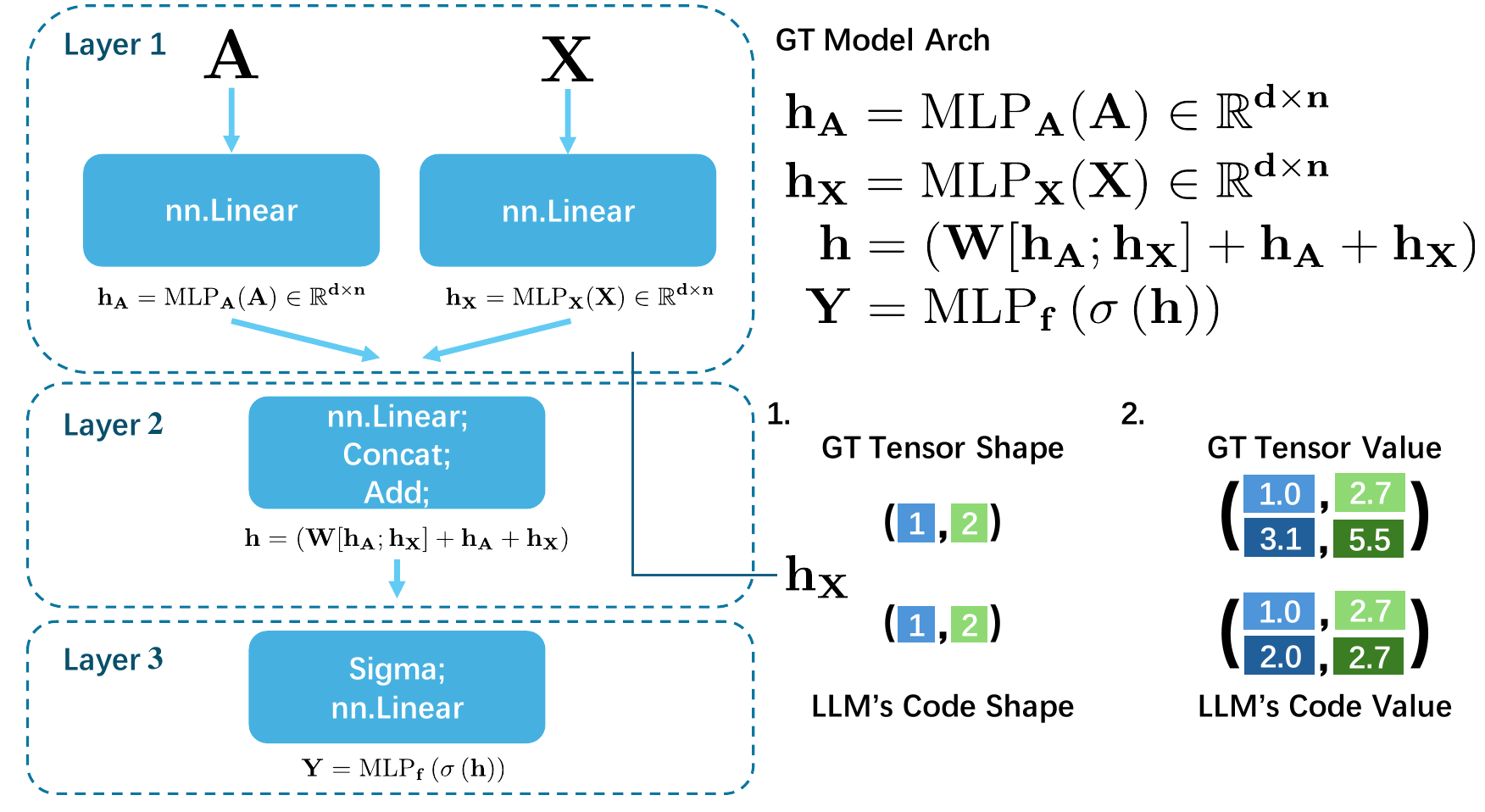}
    \caption{An example of metrics calculation for model architecture implementation task.}
    \label{fig:model_implementation_example}
\end{figure}

An example of the calculation is shown in Figure~\ref{fig:model_implementation_example}, using model LinkX ~\citep{linkx} as an example. Meanwhile, we also include the ``average running success rate'' as the basic metric for the model architecture implementation task, which is the same as the formula implementation task.


\section{Experiments}

\subsection{Experimental Settings}
As we have numeric input and output in R\&D tasks, we set numeric equability with 1e-6 as tolerance for the evaluation of the implementation of methods. We set the base models as GPT-4-turbo~\citep{GPT4Report}, GPT-4-32k~\citep{GPT4Report}, GPT-35-turbo-16k~\citep{GPT4Report} and Llama2~\citep{Llama2} for the experiments. All the methods mentioned above, and their corresponding results are executed with Azure OpenAI API. There is no external data, resources, human feedback, or internet access involved in the experiments. We perform 20 independent attempts for each model and calculate the average and maximum value of each metric. As most of our input data is encoded in the form of document files, we first use parsing tools to extract text content from files. Azure document intelligence API (4.0) is used for parsing reports and academic papers in PDF format.

\subsection{Results of Method Extraction}


We evaluate the information extraction ability of models. As shown in Table~\ref{exp:extraction}, we observe that GPT-4-turbo, GPT-4-32k, LLaMa3-70b, and LLaMa2-70b achieve competitive performance, which makes them possible to perform information extraction automatically. The performance of the four foundation models is stronger than of Phi3-4k, indicating more future endeavors to improve the extraction ability of Phi3-4k.

\begin{wraptable}{r}{6cm}
\centering
\scalebox{0.8}{
\begin{tabular}{lccc}
\toprule
\textbf{Metrics}  & \textbf{Precion} & \textbf{Recall} & \textbf{F1} \\ \hline
{GPT-4-turbo}                            & {0.818}     & {0.800}           & {0.809}\\
{GPT-4-32k}                             & {0.818}     & {0.818}           & {0.818}\\
{LLaMa3-70b}                           & \textbf{0.909}   & {0.833}             & \textbf{0.869}\\ 
{LLaMa2-70b}                             & {0.818}     & \textbf{0.900}           & {0.857}\\
{Phi3-4k}                              & {0.636}      & {0.750}           & {0.688}\\ \bottomrule
\end{tabular}}
\caption{Results of method extraction.}
\label{exp:extraction}
\end{wraptable}

\subsection{Results of Formula Implementation}
In this section, we compare the performance of different models in the model architecture implementation task. We use the proposed metrics to evaluate the performance of the models. The results are shown in Table~\ref{exp:facimp-gpt4turbo} and Table~\ref{exp:facimp-overall}. We observe that the GPT-4-turbo achieves better performance than GPT-35-turbo and Phi3-128k in the model architecture implementation task. Overall experimental results indicate ample room for further research on the difficulty of the task and the challenges in automating R\&D tasks. Specifically, we obtain the following four major findings revealed by the experimental results.

\begin{table}[htbp]
  \centering
  \scalebox{0.8}{
  \begin{tabular}{@{}lllcccc@{}}
    \toprule
    Data & Difficulty & Formula & \multicolumn{1}{c}{Avg. Exec.} & \multicolumn{1}{c}{Avg. Format} & \multicolumn{1}{c}{Avg. Corr.} & \multicolumn{1}{c}{Max. Corr.} \\
    \midrule
    \multirow{9}{*}{Data I} & \multirow{3}{*}{Easy} & PB\_ROE & 0.650 & 0.050 & 0.852 & 0.852 \\
     &  & PB\_ROE\_2 & 0.600 & 0.200 & 0.875 & 1.000 \\
     &  & PB\_ROE\_3 & 0.600 & 0.300 & 0.726 & 1.000 \\
    \cmidrule{2-7}
     & \multirow{3}{*}{Medium} & ROE\_movement & 0.950 & 0.750 & 0.934 & 1.000 \\
     &  & ROE\_movement\_10 & 0.900 & 0.800 & 0.803 & 1.000 \\
     &  & ROE\_movement\_20 & 0.950 & 0.750 & 0.703 & 1.000 \\
    \cmidrule{2-7}
     & \multirow{3}{*}{Hard} & PB\_ROE\_movement & 0.600 & 0.450 & 0.516 & 0.897 \\
     &  & PB\_ROE\_movement\_10 & 0.650 & 0.300 & 0.327 & 0.896 \\
     &  & PB\_ROE\_movement\_20 & 0.550 & 0.500 & 0.244 & 0.896 \\
    \midrule
    \multirow{9}{*}{Data II} & \multirow{3}{*}{Easy} & mid\_price & 0.800 & 0.100 & 1.000 & 1.000 \\
     &  & mid\_price\_2 & 0.850 & 0.000 & NaN & NaN \\
     &  & mid\_price\_3 & 0.850 & 0.000 & NaN & NaN \\
    \cmidrule{2-7}
     & \multirow{3}{*}{Medium} & liquidity\_imbalance & 0.500 & 0.050 & 1.000 & 1.000 \\
     &  & liquidity\_imbalance\_2 & 0.900 & 0.150 & 0.694 & 1.000 \\
     &  & liquidity\_imbalance\_3 & 0.450 & 0.100 & 1.000 & 1.000 \\
    \cmidrule{2-7}
     & \multirow{3}{*}{Hard} & micro\_price & 0.850 & 0.000 & NaN & NaN \\
     &  & micro\_price\_2 & 0.600 & 0.000 & NaN & NaN \\
     &  & micro\_price\_3 & 0.600 & 0.100 & 1.000 & 1.000 \\
    \midrule
    \multirow{9}{*}{Data III} & \multirow{3}{*}{Easy} & alpha053 & 0.950 & 0.700 & 0.933 & 1.000 \\
     &  & alpha053\_15 & 0.950 & 0.650 & 0.872 & 1.000 \\
     &  & alpha053\_5 & 1.000 & 0.650 & 0.676 & 1.000 \\
    \cmidrule{2-7}
     & \multirow{3}{*}{Medium} & alpha\_pv\_diff & 1.000 & 0.600 & 0.513 & 1.000 \\
     &  & alpha\_pv\_diff\_15 & 0.950 & 0.750 & 0.258 & 1.000 \\
     &  & alpha\_pv\_diff\_20 & 1.000 & 0.750 & 0.441 & 1.000 \\
    \cmidrule{2-7}
     & \multirow{3}{*}{Hard} & alpha\_pv\_diff\_pct & 0.950 & 0.700 & 0.375 & 1.000 \\
     &  & alpha\_pv\_diff\_pct\_15 & 0.900 & 0.450 & 0.236 & 1.000 \\
     &  & alpha\_pv\_diff\_pct\_20 & 1.000 & 0.350 & 0.358 & 1.000 \\
    \midrule
    \multirow{4}{*}{Overall} & \multirow{4}{*}{N/A} & Avg. Data I & 0.717 & 0.456 & 0.665 & 0.949 \\
     &  & Avg. Data II & 0.711 & 0.056 & 0.522 & 0.556 \\
     &  & Avg. Data III & 0.967 & 0.622 & 0.518 & 1.000 \\
     &  & Mean Value & 0.798 & 0.378 & 0.568 & 0.835 \\
    \bottomrule
  \end{tabular}}
  \caption{The performance of GPT-4-turbo in formula implementation.}
  \label{exp:facimp-gpt4turbo}
\end{table}

\noindent \textbf{LLM agents hold promising potential to tackle D-CARD.}
We can observe from Table~\ref{exp:facimp-gpt4turbo} and Table~\ref{exp:facimp-overall} that GPT-4 possesses the ability to tackle some simple D-CARD cases without adopting any additional techniques. Specifically, GPT-4 achieves a high maximum correlation coefficient with the ground-truth results in implementing both easy and medium formulations: GPT-4-turbo achieves the maximum correlation value in implementing easy and medium formulas. However, GPT-4 fails to precisely match the exact ground-truth values due to some minor mistakes, such as missing the domain common knowledge (e.g., using percent change rather than difference when calculating growth), mismatching the output format, and unnecessarily introducing additional computational operations.

\noindent \textbf{Precisely understanding and selecting data requires more detailed data information in D-CARD.}
As shown in Table~\ref{exp:facimp-gpt4turbo}, we observe a special situation where GPT-4 significantly fails to implement a simple formulation while succeeding in implementing the harder ones. After analyzing its generated code, we find that GPT-4 confuses the different semantic meanings of data features due to their close natural language descriptions, which renders the subsequent calculation ineffective. For example, GPT-4 confuses the two terms named ``volume'' and ``volatility'' and always opts to use ``volume'' data when ``volatility'' is required. If we manually improve our initial prompt by adding a more detailed description, GPT-4 succeeds in understanding the semantic difference and obtains over 99\% performance in the accuracy of values.

\noindent \textbf{The ability to query domain-specific knowledge is a basic requirement of D-CARD methods.}
As we mentioned in the first finding, missing domain common knowledge impedes GPT-4 from calculating precisely matched final results. Additionally, we find that the implementation of some operations in a formulation also requires domain-specific knowledge. For example, in the financial domain, it's clear enough for financial practitioners to implement the operation named ``IndNeutralize(x,g)'' by merely giving the description ``x cross-sectionally neutralized against groups g''. However, in the code generated by GPT-4, it defines a function named ``IndNeutralize(series, industry)'' and leaves its content blank by merely adding a notation ``Please replace this with your actual function definition''.

\noindent \textbf{The more complex the method is, the more unstable the model performance is.}
As shown in the columns of Table~\ref{exp:facimp-gpt4turbo} named ``avg. exec.'', ``avg. form.'', and ``avg. corr.'', respectively, we can observe that the performance variance of GPT-4 is significantly higher as the complexity of formulations increases. In 20 times of execution, GPT-4 generates the successfully executed code 18 times when implementing the medium mid\_price while only three times in implementing hard alpha\_pv.

\begin{table}[!h]
  \centering
  \scalebox{0.8}{%
    \begin{tabular}{lllcccc}
      \toprule
      & & & \multicolumn{1}{c}{avg. exec.} & \multicolumn{1}{c}{avg. format} & \multicolumn{1}{c}{avg. corr.} & \multicolumn{1}{c}{max. corr.} \\
      \midrule
      \multirow{4}{*}{GPT-4—turbo} & & Data I & 0.717 & 0.456 & 0.665 & 0.949 \\
       & & Data II & 0.711 & 0.056 & 0.522 & 0.556 \\
       & & Data III & 0.967 & 0.622 & 0.518 & 1.000 \\
       & & Mean Value & 0.798 & 0.378 & 0.568 & 0.835 \\
      \midrule
      \multirow{4}{*}{GPT-35-turbo} & & Data I & 0.556 & 0.100 & 0.323 & 0.453 \\
       & & Data II & 0.567 & 0.000 & 0.000 & 0.000 \\
       & & Data III & 0.767 & 0.389 & 0.431 & 0.696 \\
       & & Mean Value & 0.630 & 0.163 & 0.251 & 0.383 \\
      \midrule
      \multirow{4}{*}{Phi3-128k} & & Data I & 0.117 & 0.111 & 0.186 & 0.222 \\
       & & Data II & 0.172 & 0.000 & 0.000 & 0.000 \\
       & & Data III & 0.056 & 0.022 & 0.063 & 0.084 \\
       & & Mean Value & 0.115 & 0.044 & 0.083 & 0.102 \\
      \bottomrule
    \end{tabular}
  }
  \caption{Comparison of our benchmark for different models}
  \label{exp:facimp-overall}
\end{table}

As shown in Table ~\ref{exp:facimp-overall}, the performance of GPT-35-turbo and Phi3-128k is poor, even failing in execution codes. However, GPT-4 models shown in Table~\ref{exp:facimp-gpt4turbo} have a much better performance. This indicates that the performance of the model in the data-centric R\&D task is highly related to the model's pre-training and capacity. Therefore, we posit that continually training and improving the foundation model is a promising direction for future research in the field of data-centric R\&D tasks.

\subsection{Results of Model Architecture Implementation}
In this section, we compare the performance of different LLMs in the model architecture implementation task and summarize the results in Table ~\ref{exp:model_arch_gpt4} and ~\ref{exp:model_arch_gpt35_llama2}. As shown in the table, we can see the GPT-4-turbo, GPT-35-turbo-16k, and GPT-4-32K have similar running success rates but differ variously in tvc. and tsc.. The LLaMa-2-70b has the lowest running success rate and other metrics. Notice that even though a significant gap still exists between GPT-35, LLaMa-2, and GPT-4, it is much smaller than the gap in the formula implementation task. The overall running success rates are also higher than the formula implementation task. We can conclude that we can have similar observations in the model architecture implementation task as in the formula implementation task.

\begin{table}[h]
    \centering
    \begin{adjustbox}{width=\textwidth, center}
        \begin{tabular}{llcccccccccc}
            \hline
            \multirow{4}{*}{\textbf{Model Name}}             &
            \multirow{4}{*}{\textbf{Difficulty}}             &
            \multicolumn{5}{c}{\multirow{2}{*}{GPT-4-turbo}} &
            \multicolumn{5}{c}{\multirow{2}{*}{GPT-4-32K}}     \\
                                                             &
                                                             &
            \multicolumn{5}{c}{}                             &
            \multicolumn{5}{c}{}                               \\ \cline{3-12}
                                                             &
                                                             &
            \multirow{2}{*}{\textbf{avg. exe.}}               &
            \multirow{2}{*}{\textbf{avg. tsc.}}                &
            \multirow{2}{*}{\textbf{avg. tvc.}}                &
            \multirow{2}{*}{\textbf{max. tsc.}}                &
            \multirow{2}{*}{\textbf{max. tvc.}}                &
            \multirow{2}{*}{\textbf{avg. exe.}}                &
            \multirow{2}{*}{\textbf{avg. tsc.}}                &
            \multirow{2}{*}{\textbf{avg. tvc.}}                &
            \multirow{2}{*}{\textbf{max. tsc.}}                &
            \multirow{2}{*}{\textbf{max. tvc.}}                  \\
                                                             &
                                                             &
                                                             &
                                                             &
                                                             &
                                                             &
                                                             &
                                                             &
                                                             &
                                                             &
                                                             &
            \\ \hline
            \multirow{2}{*}{PMLP}                            &
            \multirow{2}{*}{Easy}                            &
            \multirow{2}{*}{100.00\%}                        &
            \multirow{2}{*}{1.00}                            &
            \multirow{2}{*}{1.00}                            &
            \multirow{2}{*}{1.00}                            &
            \multirow{2}{*}{1.00}                            &
            \multirow{2}{*}{100.00\%}                        &
            \multirow{2}{*}{1.00}                            &
            \multirow{2}{*}{1.00}                            &
            \multirow{2}{*}{1.00}                            &
            \multirow{2}{*}{1.00}                              \\
                                                             &
                                                             &
                                                             &
                                                             &
                                                             &
                                                             &
                                                             &
                                                             &
                                                             &
                                                             &
                                                             &
            \\
            \multirow{2}{*}{LinkX}                           &
            \multirow{2}{*}{Easy}                            &
            \multirow{2}{*}{100.00\%}                        &
            \multirow{2}{*}{1.00}                            &
            \multirow{2}{*}{0.85}                            &
            \multirow{2}{*}{1.00}                            &
            \multirow{2}{*}{1.00}                            &
            \multirow{2}{*}{100.00\%}                        &
            \multirow{2}{*}{0.90}                            &
            \multirow{2}{*}{0.90}                            &
            \multirow{2}{*}{1.00}                            &
            \multirow{2}{*}{1.00}                              \\
                                                             &
                                                             &
                                                             &
                                                             &
                                                             &
                                                             &
                                                             &
                                                             &
                                                             &
                                                             &
                                                             &
            \\
            \multirow{2}{*}{VisNet}                          &
            \multirow{2}{*}{Hard}                            &
            \multirow{2}{*}{45.00\%}                         &
            \multirow{2}{*}{0.29}                            &
            \multirow{2}{*}{0.09}                            &
            \multirow{2}{*}{0.37}                            &
            \multirow{2}{*}{0.49}                            &
            \multirow{2}{*}{45.00\%}                         &
            \multirow{2}{*}{0.21}                            &
            \multirow{2}{*}{0.09}                            &
            \multirow{2}{*}{0.37}                            &
            \multirow{2}{*}{0.49}                              \\
                                                             &
                                                             &
                                                             &
                                                             &
                                                             &
                                                             &
                                                             &
                                                             &
                                                             &
                                                             &
                                                             &
            \\
            \multirow{2}{*}{AntiSymmetric}                   &
            \multirow{2}{*}{Medium}                          &
            \multirow{2}{*}{80.00\%}                         &
            \multirow{2}{*}{0.71}                            &
            \multirow{2}{*}{0.59}                            &
            \multirow{2}{*}{0.73}                            &
            \multirow{2}{*}{0.88}                            &
            \multirow{2}{*}{70.00\%}                         &
            \multirow{2}{*}{0.56}                            &
            \multirow{2}{*}{0.66}                            &
            \multirow{2}{*}{0.66}                            &
            \multirow{2}{*}{0.88}                              \\
                                                             &
                                                             &
                                                             &
                                                             &
                                                             &
                                                             &
                                                             &
                                                             &
                                                             &
                                                             &
                                                             &
            \\
            \multirow{2}{*}{GPSConv}                         &
            \multirow{2}{*}{Medium}                          &
            \multirow{2}{*}{75.00\%}                         &
            \multirow{2}{*}{0.56}                            &
            \multirow{2}{*}{0.62}                            &
            \multirow{2}{*}{0.65}                            &
            \multirow{2}{*}{1.00}                            &
            \multirow{2}{*}{75.00\%}                         &
            \multirow{2}{*}{0.53}                            &
            \multirow{2}{*}{0.62}                            &
            \multirow{2}{*}{0.65}                            &
            \multirow{2}{*}{0.72}                              \\
                                                             &
                                                             &
                                                             &
                                                             &
                                                             &
                                                             &
                                                             &
                                                             &
                                                             &
                                                             &
                                                             &
            \\
            \multirow{2}{*}{DirGNNConv}                      &
            \multirow{2}{*}{Medium}                          &
            \multirow{2}{*}{100.00\%}                        &
            \multirow{2}{*}{0.80}                            &
            \multirow{2}{*}{0.68}                            &
            \multirow{2}{*}{0.86}                            &
            \multirow{2}{*}{0.94}                            &
            \multirow{2}{*}{90.00\%}                         &
            \multirow{2}{*}{0.65}                            &
            \multirow{2}{*}{0.62}                            &
            \multirow{2}{*}{0.82}                            &
            \multirow{2}{*}{0.91}                              \\
                                                             &
                                                             &
                                                             &
                                                             &
                                                             &
                                                             &
                                                             &
                                                             &
                                                             &
                                                             &
                                                             &
            \\ \hline
        \end{tabular}
    \end{adjustbox}
    \caption{The performance of GPT-4-turbo and GPT-4-32k on model architecture implementation tasks.}
    \label{exp:model_arch_gpt4}
\end{table}

\begin{table}[h]
    \centering
    \begin{adjustbox}{width=\textwidth, center}
\begin{tabular}{llccccccccccccccc}
\hline
\multirow{4}{*}{\textbf{Model Name}} & \multirow{4}{*}{\textbf{Difficulty}} & \multicolumn{5}{c}{\multirow{2}{*}{GPT-35-turbo-16k}}                                                                                                                                       & \multicolumn{5}{c}{\multirow{2}{*}{LLaMa-2-70b}}                                                                                                                                            & \multicolumn{5}{c}{\multirow{2}{*}{LLaMa-3-70b}}                                                                                                                                            \\
                                     &                                      & \multicolumn{5}{c}{}                                                                                                                                                                        & \multicolumn{5}{c}{}                                                                                                                                                                        & \multicolumn{5}{c}{}                                                                                                                                                                        \\ \cline{3-17} 
                                     &                                      & \multirow{2}{*}{\textbf{avg. exe.}} & \multirow{2}{*}{\textbf{avg. tsc.}} & \multirow{2}{*}{\textbf{avg. tvc.}} & \multirow{2}{*}{\textbf{max. tsc.}} & \multirow{2}{*}{\textbf{max. tvc.}} & \multirow{2}{*}{\textbf{avg. exe.}} & \multirow{2}{*}{\textbf{avg. tsc.}} & \multirow{2}{*}{\textbf{avg. tvc.}} & \multirow{2}{*}{\textbf{max. tsc.}} & \multirow{2}{*}{\textbf{max. tvc.}} & \multirow{2}{*}{\textbf{avg. exe.}} & \multirow{2}{*}{\textbf{avg. tsc.}} & \multirow{2}{*}{\textbf{avg. tvc.}} & \multirow{2}{*}{\textbf{max. tsc.}} & \multirow{2}{*}{\textbf{max. tvc.}} \\
                                     &                                      &                                     &                                     &                                     &                                     &                                     &                                     &                                     &                                     &                                     &                                     &                                     &                                     &                                     &                                     &                                     \\ \hline
\multirow{2}{*}{PMLP}                & \multirow{2}{*}{Easy}                & \multirow{2}{*}{100.00\%}           & \multirow{2}{*}{0.75}               & \multirow{2}{*}{0.75}               & \multirow{2}{*}{1.00}               & \multirow{2}{*}{1.00}               & \multirow{2}{*}{60.00\%}            & \multirow{2}{*}{0.45}               & \multirow{2}{*}{0.55}               & \multirow{2}{*}{1.00}               & \multirow{2}{*}{1.00}               & \multirow{2}{*}{100.00\%}           & \multirow{2}{*}{0.85}               & \multirow{2}{*}{0.75}               & \multirow{2}{*}{1.00}               & \multirow{2}{*}{1.00}               \\
                                     &                                      &                                     &                                     &                                     &                                     &                                     &                                     &                                     &                                     &                                     &                                     &                                     &                                     &                                     &                                     &                                     \\
\multirow{2}{*}{LinkX}               & \multirow{2}{*}{Easy}                & \multirow{2}{*}{100.00\%}           & \multirow{2}{*}{0.60}               & \multirow{2}{*}{0.34}               & \multirow{2}{*}{1.00}               & \multirow{2}{*}{1.00}               & \multirow{2}{*}{30.00\%}            & \multirow{2}{*}{0.20}               & \multirow{2}{*}{0.15}               & \multirow{2}{*}{1.00}               & \multirow{2}{*}{1.00}               & \multirow{2}{*}{100.00\%}           & \multirow{2}{*}{0.80}               & \multirow{2}{*}{0.65}               & \multirow{2}{*}{1.00}               & \multirow{2}{*}{1.00}               \\
                                     &                                      &                                     &                                     &                                     &                                     &                                     &                                     &                                     &                                     &                                     &                                     &                                     &                                     &                                     &                                     &                                     \\
\multirow{2}{*}{VisNet}              & \multirow{2}{*}{Hard}                & \multirow{2}{*}{5.00\%}             & \multirow{2}{*}{0.03}               & \multirow{2}{*}{0.00}               & \multirow{2}{*}{0.16}               & \multirow{2}{*}{0.40}               & \multirow{2}{*}{0.00\%}             & \multirow{2}{*}{0.00}               & \multirow{2}{*}{0.00}               & \multirow{2}{*}{0.00}               & \multirow{2}{*}{0.00}               & \multirow{2}{*}{40\%}               & \multirow{2}{*}{0.27}               & \multirow{2}{*}{0.24}               & \multirow{2}{*}{0.33}               & \multirow{2}{*}{0.42}               \\
                                     &                                      &                                     &                                     &                                     &                                     &                                     &                                     &                                     &                                     &                                     &                                     &                                     &                                     &                                     &                                     &                                     \\
\multirow{2}{*}{AntiSymmetric}       & \multirow{2}{*}{Medium}              & \multirow{2}{*}{45.00\%}            & \multirow{2}{*}{0.16}               & \multirow{2}{*}{0.21}               & \multirow{2}{*}{0.61}               & \multirow{2}{*}{0.22}               & \multirow{2}{*}{0.00\%}             & \multirow{2}{*}{0.00}               & \multirow{2}{*}{0.00}               & \multirow{2}{*}{0.00}               & \multirow{2}{*}{0.00}               & \multirow{2}{*}{80\%}               & \multirow{2}{*}{0.62}               & \multirow{2}{*}{0.70}               & \multirow{2}{*}{0.71}               & \multirow{2}{*}{0.88}               \\
                                     &                                      &                                     &                                     &                                     &                                     &                                     &                                     &                                     &                                     &                                     &                                     &                                     &                                     &                                     &                                     &                                     \\
\multirow{2}{*}{GPSConv}             & \multirow{2}{*}{Medium}              & \multirow{2}{*}{45.00\%}            & \multirow{2}{*}{0.24}               & \multirow{2}{*}{0.19}               & \multirow{2}{*}{0.45}               & \multirow{2}{*}{0.42}               & \multirow{2}{*}{0.00\%}             & \multirow{2}{*}{0.00}               & \multirow{2}{*}{0.00}               & \multirow{2}{*}{0.00}               & \multirow{2}{*}{0.00}               & \multirow{2}{*}{75\%}               & \multirow{2}{*}{0.51}               & \multirow{2}{*}{0.59}               & \multirow{2}{*}{0.65}               & \multirow{2}{*}{1.00}               \\
                                     &                                      &                                     &                                     &                                     &                                     &                                     &                                     &                                     &                                     &                                     &                                     &                                     &                                     &                                     &                                     &                                     \\
\multirow{2}{*}{DirGNNConv}          & \multirow{2}{*}{Medium}              & \multirow{2}{*}{65.00\%}            & \multirow{2}{*}{0.56}               & \multirow{2}{*}{0.29}               & \multirow{2}{*}{0.71}               & \multirow{2}{*}{0.42}               & \multirow{2}{*}{0.00\%}             & \multirow{2}{*}{0.00}               & \multirow{2}{*}{0.00}               & \multirow{2}{*}{0.00}               & \multirow{2}{*}{0.00}               & \multirow{2}{*}{90\%}               & \multirow{2}{*}{0.72}               & \multirow{2}{*}{0.68}               & \multirow{2}{*}{0.84}               & \multirow{2}{*}{0.91}               \\
                                     &                                      &                                     &                                     &                                     &                                     &                                     &                                     &                                     &                                     &                                     &                                     &                                     &                                     &                                     &                                     &                                     \\ \hline
\end{tabular}
    \end{adjustbox}
    \caption{The performance of GPT-35-turbo-16k, LLaMa-2-70b and LLaMa-3-70b on model architecture implementation tasks.}
    \label{exp:model_arch_gpt35_llama2}
\end{table}




\section{Limitation}
The $\text{RD}^2$Bench framework, while innovative, only evaluates the most representative base LLM, such as GPT4, LLama3, LLama2, GPT35, without further evaluate more open source models. Meanwhile, this paper only include most representative R\&D domains and problems and focus on data driven scenarios, which can be extended more in the future to show its generalizability.
For more details and a more comprehensive evaluation, we will include them in not-too-distant future works. We believe that the benchmark will be a valuable tool for the community to evaluate the performance of the models in the data-centric R\&D tasks and to develop new models and techniques to address the challenges and opportunities in the domain.

\section{Conclusion}
In this paper, we serve as the first effort to tackle the real-world data-centric automatic R\&D scenario in the hope of significantly improving the research efficiency of scientists and thus contributing to the revolution of human productivity. Specifically, we first propose $\text{RD}^2$Bench that benchmarks all the operations in D-CARD as a whole to navigate future work toward the ultimate goal of automating data-centric R\&D directly. $\text{RD}^2$Bench focuses on evaluating the interaction and synergistic effects of various model capabilities and aiding in selecting the well-performing trustworthy models. Based on $\text{RD}^2$Bench, we find that although the most SOTA GPT-4 shows its promising potency in tackling D-CARD, there remains ample room for future work.

\bibliography{main}
\bibliographystyle{plain}


\appendix

\section{Formula implementation task metrics calculation details}
\label{App:A}
As mentioned above, we have multiple metrics (the average and maxima score across multiple independent attempts, including ”running success rate”, ”format success rate”, ”pearson correlation” and ”value accuracy”). Assume the ground truth factor value is $\mathbf{Y}$ with length $n$ (the length of the time series), and the generated factor value is $\mathbf{Y}^{*}$, the calculation of the metrics is as follows:

\textbf{Running success} is defined as successful execution. Any error that occurs in the Python interpreter during the execution that stops the execution is considered a failure. We calculate the ratio of the number of successful execution times to the total number of attempts, denoted as avg. exe.

\textbf{Pearson correlation} is the correlation between the ground truth factor value and the generated factor value.
\begin{equation*}
    \text{corr.} = \frac{\sum_{i=1}^{n}(\mathbf{Y}^{*}_{i} - \bar{\mathbf{Y}^{*}})(\mathbf{Y}_{i} - \bar{\mathbf{Y}})}{\sqrt{\sum_{i=1}^{n}(\mathbf{Y}^{*}_{i} - \bar{\mathbf{Y}^{*}})^{2}}\sqrt{\sum_{i=1}^{n}(\mathbf{Y}_{i} - \bar{\mathbf{Y}})^{2}}},
\end{equation*}

\textbf{Format success} is defined as successful format matching, which means the final output dataframe format is (datetime, factor\_name). We calculate the ratio of the number of matched result formats to the total number of attempts, denoted as avg. form.

\textbf{Value accuracy} is the accuracy of the generated factor value, which can be formulated as:
\begin{equation*}
    \text{acc.} = \frac{1}{n}\sum_{i=1}^{n}\mathbb{I}(|\mathbf{Y}^{*}_{i} - \mathbf{Y}_{i}| < t),
\end{equation*}

Please note that we set the tolerance $t$ for the value accuracy as 1e-6 in this paper, which means two values are considered as equal if the absolute difference is less than 1e-6.

\section{Data collection details}
\label{App:B}
As mentioned in the previous section, we collected papers including ~\citep{antisymmetric,DirGNN,GPSConv,linkx,pmlp,Visnet} and corresponding codes using pyg~\citep{pyg}, which are listed in the following table.

\begin{table}[h]
    \centering
    \begin{tabular}{cccc}
        \hline
        Paper                                                            & Type  & Difficulty & GT Code                                                                                                              \\ \hline
        \href{https://arxiv.org/abs/2212.09034}{PMLP}                    & Model & Easy       & \href{https://github.com/pyg-team/pytorch_geometric/blob/master/torch_geometric/nn/models/pmlp.py}{Link}             \\
        \href{https://arxiv.org/abs/2110.14446}{LinkX}                   & Model & Easy       & \href{https://github.com/pyg-team/pytorch_geometric/blob/master/torch_geometric/nn/models/linkx.py}{Link}            \\
        \href{https://openreview.net/forum?id=J3Y7cgZOOS}{AntiSymmetric} & Layer & Medium     & \href{https://github.com/pyg-team/pytorch_geometric/blob/master/torch_geometric/nn/conv/antisymmetric_conv.py}{Link} \\
        \href{https://arxiv.org/abs/2205.12454}{GPSConv}                 & Layer & Medium     & \href{ https://github.com/pyg-team/pytorch_geometric/blob/master/torch_geometric/nn/conv/gps_conv.py}{Link}          \\
        \href{https://arxiv.org/abs/2305.10498}{DirGNNCOnv}              & Layer & Medium     & \href{https://github.com/pyg-team/pytorch_geometric/blob/master/torch_geometric/nn/conv/dir_gnn_conv.py}{Link}       \\
        \href{https://arxiv.org/abs/2210.16518}{VisNet}                  & Model & Hard       & \href{https://github.com/pyg-team/pytorch_geometric/blob/master/torch_geometric/nn/models/visnet.py}{Link}           \\ \hline
    \end{tabular}
    \caption{Papers and corresponding ground truth implementation codes for the model architecture implementation task}
\end{table}
\section{Broader Impact}
The proposed $\text{RD}^2$Bench has the potential to significantly impact the scientific community and industries reliant on R\&D. By automating the tedious aspects of R\&D, researchers can focus on more creative and innovative aspects of their work, potentially accelerating the pace of discoveries. Smaller institutions or individual researchers with limited resources might benefit from automated tools that reduce the need for extensive human labor, making high-level R\&D more accessible. Automation of R\&D can reduce costs and time-to-market for new technologies, fostering faster economic growth and competitiveness

\begin{table}[!htbp]
  \centering
  \adjustbox{max width=\textwidth}{%
  \begin{tabular}{@{}lllcccc@{}}
    \toprule
    & & & \multicolumn{1}{c}{avg. exec.} & \multicolumn{1}{c}{avg. format} & \multicolumn{1}{c}{avg. corr.} & \multicolumn{1}{c}{max. corr.} \\
    \midrule
    \multirow{9}{*}{Fundamental} & \multirow{3}{*}{Easy} & PB\_ROE & 0.400 & 0.000 & NaN & NaN \\
     &  & PB\_ROE\_2 & 0.600 & 0.000 & NaN & NaN \\
     &  & PB\_ROE\_3 & 0.600 & 0.200 & 0.521 & 0.999 \\
    \cmidrule{2-7}
     & \multirow{3}{*}{Medium} & ROE\_movement & 0.800 & 0.300 & 0.339 & 1.000 \\
     &  & ROE\_movement\_10 & 0.600 & 0.100 & 1.000 & 1.000 \\
     &  & ROE\_movement\_20 & 0.900 & 0.200 & 0.967 & 1.000 \\
    \cmidrule{2-7}
     & \multirow{3}{*}{Hard} & PB\_ROE\_movement & 0.200 & 0.100 & 0.078 & 0.078 \\
     &  & PB\_ROE\_movement\_10 & 0.500 & 0.000 & NaN & NaN \\
     &  & PB\_ROE\_movement\_20 & 0.400 & 0.000 & NaN & NaN \\
    \midrule
    \multirow{9}{*}{High Frequency} & \multirow{3}{*}{Easy} & mid\_price & 0.600 & 0.000 & NaN & NaN \\
     &  & mid\_price\_2 & 0.500 & 0.000 & NaN & NaN \\
     &  & mid\_price\_3 & 0.600 & 0.000 & NaN & NaN \\
    \cmidrule{2-7}
     & \multirow{3}{*}{Medium} & liquidity\_imbalance & 0.200 & 0.000 & NaN & NaN \\
     &  & liquidity\_imbalance\_2 & 0.800 & 0.000 & NaN & NaN \\
     &  & liquidity\_imbalance\_3 & 0.500 & 0.000 & NaN & NaN \\
    \cmidrule{2-7}
     & \multirow{3}{*}{Hard} & micro\_price & 0.400 & 0.000 & NaN & NaN \\
     &  & micro\_price\_2 & 0.700 & 0.000 & NaN & NaN \\
     &  & micro\_price\_3 & 0.800 & 0.000 & NaN & NaN \\
    \midrule
    \multirow{9}{*}{Price Volume} & \multirow{3}{*}{Easy} & alpha053 & 0.800 & 0.500 & 0.809 & 1.000 \\
     &  & alpha053\_15 & 0.700 & 0.500 & 0.806 & 1.000 \\
     &  & alpha053\_5 & 0.700 & 0.500 & 0.440 & 1.000 \\
    \cmidrule{2-7}
     & \multirow{3}{*}{Medium} & alpha\_pv\_diff & 0.800 & 0.700 & 0.304 & 1.000 \\
     &  & alpha\_pv\_diff\_15 & 0.700 & 0.400 & 0.259 & 1.000 \\
     &  & alpha\_pv\_diff\_20 & 0.600 & 0.400 & 1.000 & 1.000 \\
    \cmidrule{2-7}
     & \multirow{3}{*}{Hard} & alpha\_pv\_diff\_pct & 0.800 & 0.200 & -0.011 & -0.011 \\
     &  & alpha\_pv\_diff\_pct\_15 & 0.900 & 0.200 & 0.096 & 0.096 \\
     &  & alpha\_pv\_diff\_pct\_20 & 0.900 & 0.100 & 0.176 & 0.176 \\
    \midrule
    \multirow{4}{*}{gpt3.5} & \multirow{4}{*}{N/A} & Fundamental Avg & 0.556 & 0.100 & 0.323 & 0.453 \\
     &  & High Frequency Avg & 0.567 & 0.000 & 0.000 & 0.000 \\
     &  & Price Volume Avg & 0.767 & 0.389 & 0.431 & 0.696 \\
     &  & mean value (0 for NaN) & 0.630 & 0.163 & 0.251 & 0.383 \\
    \bottomrule
  \end{tabular}
  }
  \caption{The performance of gpt-35-turbo in formula implementation.}
\end{table}

\begin{table}[!htbp]
  \centering
  \adjustbox{max width=\textwidth}{%
  \begin{tabular}{@{}lllcccc@{}}
    \toprule
    & & & \multicolumn{1}{c}{avg. exec.} & \multicolumn{1}{c}{avg. format} & \multicolumn{1}{c}{avg. corr.} & \multicolumn{1}{c}{max. corr.} \\
    \midrule
    \multirow{9}{*}{Fundamental} & \multirow{3}{*}{Easy} & PB\_ROE & 0.000 & 0.000 & NaN & NaN \\
     &  & PB\_ROE\_2 & 0.050 & 0.000 & NaN & NaN \\
     &  & PB\_ROE\_3 & 0.000 & 0.000 & NaN & NaN \\
    \cmidrule{2-7}
     & \multirow{3}{*}{Medium} & ROE\_movement & 0.350 & 0.350 & 1.000 & 1.000 \\
     &  & ROE\_movement\_10 & 0.350 & 0.350 & 0.675 & 1.000 \\
     &  & ROE\_movement\_20 & 0.300 & 0.300 & NaN & NaN \\
    \cmidrule{2-7}
     & \multirow{3}{*}{Hard} oo& PB\_ROE\_movement & 0.000 & 0.000 & NaN & NaN \\
     &  & PB\_ROE\_movement\_10 & 0.000 & 0.000 & NaN & NaN \\
     &  & PB\_ROE\_movement\_20 & 0.000 & 0.000 & NaN & NaN \\
    \midrule
    \multirow{9}{*}{High Frequency} & \multirow{3}{*}{Easy} & mid\_price & 0.250 & 0.000 & NaN & NaN \\
     &  & mid\_price\_2 & 0.250 & 0.000 & NaN & NaN \\
     &  & mid\_price\_3 & 0.400 & 0.000 & NaN & NaN \\
    \cmidrule{2-7}
     & \multirow{3}{*}{Medium} & liquidity\_imbalance & 0.050 & 0.000 & NaN & NaN \\
     &  & liquidity\_imbalance\_2 & 0.150 & 0.000 & NaN & NaN \\
     &  & liquidity\_imbalance\_3 & 0.450 & 0.000 & NaN & NaN \\
    \cmidrule{2-7}
     & \multirow{3}{*}{Hard} & micro\_price & 0.000 & 0.000 & NaN & NaN \\
     &  & micro\_price\_2 & 0.000 & 0.000 & NaN & NaN \\
     &  & micro\_price\_3 & 0.000 & 0.000 & NaN & NaN \\
    \midrule
    \multirow{9}{*}{Price Volume} & \multirow{3}{*}{Easy} & alpha053 & 0.050 & 0.000 & NaN & NaN \\
     &  & alpha053\_15 & 0.000 & 0.000 & NaN & NaN \\
     &  & alpha053\_5 & 0.050 & 0.000 & NaN & NaN \\
    \cmidrule{2-7}
     & \multirow{3}{*}{Medium} & alpha\_pv\_diff & 0.250 & 0.150 & 0.413 & 0.602 \\
     &  & alpha\_pv\_diff\_15 & 0.050 & 0.000 & NaN & NaN \\
     &  & alpha\_pv\_diff\_20 & 0.000 & 0.000 & NaN & NaN \\
    \cmidrule{2-7}
     & \multirow{3}{*}{Hard} & alpha\_pv\_diff\_pct & 0.050 & 0.050 & 0.153 & 0.153 \\
     &  & alpha\_pv\_diff\_pct\_15 & 0.000 & 0.000 & NaN & NaN \\
     &  & alpha\_pv\_diff\_pct\_20 & 0.050 & 0.000 & NaN & NaN \\
    \midrule
    \multirow{4}{*}{phi3\_128k} & \multirow{4}{*}{N/A} & Fundamental Avg & 0.117 & 0.111 & 0.186 & 0.222 \\
     &  & High Frequency Avg & 0.172 & 0.000 & 0.000 & 0.000 \\
     &  & Price Volume Avg & 0.056 & 0.022 & 0.063 & 0.084 \\
     &  & mean value (0 for NaN) & 0.115 & 0.044 & 0.083 & 0.102 \\
    \bottomrule
  \end{tabular}
  }
  \caption{The performance of Phi3-128k in formula implementation.}
\end{table}

\section{Prompts}
\label{APP:C}
The prompt for the model architecture implementation task is as follows:
\begin{lstlisting}
The user is trying to implement some factors in quant investment, and you are the one to help write the Python code.
The user will provide the source data in HDF5(H5) format which you can load using pandas.read_hdf. The file is located near your Python code file which you can read from "./source_data.h5". After that, you will get a pandas dataframe with the following format:
open,close,high,low,volume,vwap,cap,IndClass.industry IndClass.sector,returns,date,instruments
2020-01-02,SH600000,158.538132,158.538132,160.699432,158.283859,4060945.0,
159.431900,647446144.0,1.0,NaN
The explanation of the example column names:
1: returns: daily close-to-close returns
2: open, close, high, low, volume: standard definitions for daily price and volume data
3: vwap: daily volume-weighted average price
4: cap: market capitalization is the total value of a company's outstanding shares of stock
5: IndClass.industry and IndClass.sector: a generic placeholder for a binary industry classification such as GICS, BICS, NAICS, SIC, etc., in indneutralize(x, IndClass.level), where level: sector, industry, etc. Multiple IndClass in the same alpha need not correspond to the same industry classification.

The user will provide you with a formulation of the factor, which contains some function calls and operators. You need to implement the function calls and operators in Python. Your code is expected to align the formulation in any form which means the user needs to get the exact factor values with your code as expected.

Your code should contain the following parts: the import part, the function part, and the main part. You should write a main function named "calculate_{function_name}" and call this function in the "if __name__ == __main__" part. Don't write any try-except block in your code. The user will catch the exception message and provide feedback to you.

User will write your code into a python file and execute the file directly with "python {your_file_name}.py". You should calculate the factor values and save the result into an HDF5(H5) file named "result.h5" in the same directory as your python file. The result file is an HDF5(H5) file containing a pandas dataframe. The index of the dataframe is the date and instrument, and the single column name is the factor name,and the value is the factor value. The result file should be saved in the same directory as your python file.

To help you write the correct code, the user might provide multiple pieces of information that help you write the correct code:
1. The user might provide you the correct code to similar factors. You should learn from these code to write the correct code.
2. The user might provide you the failed former code and the corresponding feedback to the code. The feedback contains to the execution, the code and the factor value. You should analyze the feedback and try to correct the latest code.
3. The user might provide you with suggestions for the latest failed code and some similar failed-to-correct pairs. Each pair contains the failed code with a similar error and the corresponding corrected version of the code. You should learn from these suggestions to write the correct code.

Please respond to the code in the following JSON format. Here is an example structure for the JSON output:
{
    "code": "The Python code as a string."
}
\end{lstlisting}

The prompt for the model architecture implementation task is as follows:

\begin{lstlisting}
The user is trying to implement some models or layers in deep learning, specifically in the graph learning area, and you are the one to help write the Python code.

Use PyTorch and PyG (torch_geometric) framework to implement it. You can assume the input will contain node feature X [num_nodes, feature_dim], edge_index [2, num_edges], edge_feature [num_edges, num_edge_features], y [num_nodes, *] when it is node-level targets or graph-level targets of shape [1, *], pos (node position matrix) [num_nodes, position_dim].

The user will provide you with a formulation of the model/layer. You need to implement it in Python.

Your code should contain the following parts: the import part, the function part, and the main part. You should write a main function named "calculate_function_name" and call this function in the "if __name__ == '__main__'" part. Don't write any try-except blocks in your code. The user will catch the exception message and provide the feedback to you.

User will write your code into a python file and execute the file directly with "python {your_file_name}.py". 

Please respond with the code in the following JSON format. Here is an example structure for the JSON output:
{
    "code": "The Python code as a string."
}
\end{lstlisting}

\newpage
  
\newpage

\end{document}